\documentclass[10pt,twocolumn,letterpaper]{article}

\usepackage{ijcb}
\usepackage{times}
\usepackage{epsfig}
\usepackage{graphicx}
\usepackage{amsmath}
\usepackage{amssymb}

\ijcbfinalcopy 

\title{Use of in-the-wild images for anomaly detection in face anti-spoofing}

\author{Latifah Abduh \& Ioannis Ivrissimtzis\\ 
Department of Computer Science \\ 
Durham University, UK\\ 
{\tt\small e-mail:\{latifah.a.abduh, ioannis.ivrissimtzis\}@durham.ac.uk}
}

\begin{document}
\maketitle

%%%%%%%%% ABSTRACT
\begin{abstract}
   The traditional approach to face anti-spoofing sees it as a binary classification problem, and binary classifiers are trained and validated on specialized anti-spoofing databases. One of the drawbacks of this approach is that, due to the variability of face spoofing attacks, environmental factors, and the typically small sample size, such classifiers do not generalize well to previously unseen databases. Anomaly detection, which approaches face anti-spoofing as a one-class classification problem, 
is emerging as an increasingly popular alternative approach. Nevertheless, in all existing work on anomaly detection for face anti-spoofing, the proposed training protocols utilize images from specialized anti-spoofing databases only, even though only common images of real faces are
needed. Here, we explore the use of in-the-wild images, and images from non-specialized face databases, to train one-class classifiers for face anti-spoofing. Employing a well-established technique, we train a convolutional autoencoder on real faces and compare the reconstruction error of the input against a threshold to classify a face image accordingly as either client or imposter. 
Our results show that the inclusion in the training set of in-the-wild images increases the discriminating power of the classifier significantly on an unseen database, as evidenced by a large increase in the value of the Area Under the Curve. In a limitation of our approach, we note that the problem of finding a suitable operating point on the unseen database remains a challenge, as evidenced by the values of the Half Total Error Rate. 
\end{abstract}

% ****************************************************************************** 
% ****************************************************************************** 
% ****************************************************************************** 

\section{Introduction}
Face liveness tests authenticate users of face recognition systems by processing input images and deciding whether they come from a human face or, for example, 
from printed photos held in front of the system's camera by an {\em imposter}. The main 
challenge for developing a robust face anti-spoofing is the number of types of 
{\em presentation attacks} the system must learn to recognize. For example, an imposter could be presented to the face recognition system a printed photo, a 
screen displaying a still image, or a screen replaying a video. A multitude of other factors, such as the quality of the printed photo, the resolution and type of the displaying screen, the illumination conditions of the scene, and the characteristics of the system's camera, may also have a significant effect on the performance of any anti-spoofing algorithm. Finally, a robust anti-spoofing 
algorithm should be able to cope with previously unseen attack methods, which were not anticipated prior to its deployment.

In this context, anomaly detection using classifiers trained on {\em client} 
images only are becoming an 
increasingly popular approach to face anti-spoofing \cite{arashloo2017anomaly}\cite{arashloo2018client}. The present 
work is motivated by the observation that training with client images only 
can also use in-the-wild face images, that is, face images found online, as well 
as face images from databases that do not specialize in face-anti-spoofing. 

To assess the merits of that approach, we first developed an anomaly detection anti-spoofing algorithm based on a Convolutional Autoencoder (ACE). Following a 
well-established methodology, the ACE is trained on client images and test images are classified as clients when their reconstruction error is below a threshold. First, we trained the ACE with client images from the 
Replay-Attack \cite{chingovska2012effectiveness} database, and tested it on the Replay-Attack and NUAA 
\cite{tan2010face} databases, creating a baseline. Next, we added into the training dataset in-the-wild images, which were semi-automatically collected from online sources. The results show that the classifier's discriminative power, 
as measured by the Area Under the Curve metric, increases markedly on the unseen NUAA, with a moderate only drop on Replay-Attack. Finally, we added to the training set images from databases that do not specialize in anti-spoofing, SCFace \cite{grgic2011scface} and CASIA-Web face \cite{yi2014learning} in particular, obtaining again similar results.

% ****************************************************************************** 
% ****************************************************************************** 
% ******************************************************************************  

The main contributions of the paper are: 
\begin{itemize}
\item An anomaly detection method for face anti-spoofing based on a convolutional autoencoder, which only requires RGB images for input.
\item We tested the proposed autoencoder on the previously unseen NUAA database, showing that it increases significantly when we add into the training set in-the-wild face images and face images from non-specialized databases. 
\end{itemize}
The main limitation of the proposed approach is revealed by the second set of tests, which compute the Equal Error Rate (EER) on the validation test and use the corresponding operating point, that is, the threshold against which the reconstruction error is compared, to compute the Half Total Error Rate (HTER) on the NUAA dataset. The high HTER values show that even after the enrichment of the training and validation sets, it is still not straightforward to compute a threshold giving satisfactory error rates on unseen databases.

\section{Background}

% Introduction to face anti-spoofing categorisation of the various methods 
In the past few years, a large number of methods have been proposed for the face presentation attack problem (PAD). Such methods can be classified into intrusive and non-intrusive types \cite{parveen2015face}, depending on their interference with the bio-metric data acquisition process. The non-intrusive methods, as the one proposed here, have received more attention in the literature. Another categorization of spoofing detection methods examines the way the classification algorithm 
handles image features. On the one hand, we have the traditional face anti-spoofing methods, which use
hand-crafted features and employ shallow machine learning, and on the other hand the deep learning 
methods.
% Traditional methods for face anti-spoofing 

Regarding the more traditional approaches to anti-spoofing, \cite{tan2010face} studied several 
hand-crafted feature / shallow classifier combinations. The features they studied included 
Differences of Gaussians, and features obtained through Logarithmic 
Total Variation smoothing, while their classifiers included Sparse 
Logistic Regression, Sparse Low Rank Bilinear Logistic Regression, 
and SVMs. In subsequent work, Local Binary Patterns (LBPs) are the 
most commonly used image features. In 
\cite{chingovska2012effectiveness}, LBPs are used against various 
presentation attacks, such as printed photographs, digital photos 
and videos.  

 % Deep learning methods for face anti-spoofing 
The above shallow methods do not always generalise well to previously unseen attacks. Deep learning is an alternative approach, which regularly outperforms more traditional approaches, since, 
in the context of such complex tasks, multi-layered methods seem better suited for the extraction of the high-level features of a dataset \cite{wen2015face}.

Convolutional Neural Networks (CNNs) in particular have achieved 
impressive results on a range of image and video classification 
tasks. One of the earliest attempts on liveness detection with CNNs 
is Yang et al. \cite{yang2014learn}, the results of which were 
improved by Atoum et al. \cite{atoum2017face} using a two-stream 
CNN-based face anti-spoofing method; the first stream extracts local
and holistic features and the second is used to estimate a dense 
depth map. Their proposed model performed well under intra-dataset 
testing mode. Xu et al. \cite{xu2015learning} proposed a method to 
extract a video's temporal elements using a deep neural network 
architecture combining LSTM units with a CNN containing two 
convolutions layers followed by max-pooling. Their model performed 
well under the intra-dataset testing mode, however, under a 
cross-database testing mode, these CNN models exhibit poor 
generalisability due to the overfitting of the training data.

\subsection{Anomaly detection in face anti-spoofing} 

In \cite{xiong2018unknown}, an anomaly detection classifier is proposed, which uses an autoencoder for feature extraction followed by a one-class SVM for classification, and their aim is 
generalization on previously unseen attacks, rather than previously unseen databases. In contrast, here we use a more complex convolutional autoencoder and do not rely on an SVM for classification. In \cite{yang2016deep} they use a sparse autoencoder
to encode high level features. In \cite{nikisins2019domain}, they use a convolutional autoencoder, like us, for feature extraction and MLPs for 
classification, however, their input data are much more complex consisting of Depth and NIR information together with the RGB. 
In \cite{jimenezdeep}, an anomaly detection approach based on the 
use of a deep metric was proposed, while in 
\cite{arashloo2018client}, subject-specific information was used 
with one-class classifiers to improve considerably the system's 
performance. In \cite{arashloo2017anomaly}, they use hand-crafted features such as LBPs and image quality metrics and then show that one-class classifiers work better than binary ones on cross-database testing mode. Finally, in \cite{nikisins2018effectiveness}, assuming that clients images have similar texture types, a Gaussian Mixture Model is used to represent textures, and again one-class classifiers are shown to outperform the binary ones. 

\subsection{Databases}

Replay-Attack\cite{chingovska2012effectiveness} is one of the most popular anti-spoofing databases and we used it to create our baseline. The NUAA Photograph Imposter Dataset \cite{tan2010face}, which we used for cross-database testing is also a very popular publicly available dataset. The NUAA were collected from 15 subjects using cheap 
webcams in three sessions on different environments and illumination conditions. 
Several attacks were simulated, the ones we test against here are based on a 
printed photo using A4 paper and a color printer. Notice that NUAA is consider 
a particularly challenging case when one is testing cross-database generalization 
\cite{yang2016deep}, and the use of webcams, in particular, makes it significantly different than 
Replay-Attack, CASIA-FASD \cite{zhang2012face}, or MSU-MFSD \cite{wen2015face}. 

% ****************************************************************************** 
% ****************************************************************************** 
% ******************************************************************************

\section{The experimental setup}

We employ a standard anomaly detection technique, based on a convolutional autoencoder, Hinton et al. 
Autoencoders are neural networks consisting of two parts. 
The {\em encoder} part of the network processes the input image and produces the {\em code}, a compressed representation of the input of, usually, a much lower dimension. The {\em decoder} reconstructs the original image from the code\cite{xia2015learning}. 

The entire autoencoder is trained with images from the single class of the one-class classification problem \cite{bhattad2018detecting}. The cost function is the reconstruction error, in our case, the Euclidean distance between the original and the encoded and then decoded reconstructed image. As the network is trained to minimise the reconstruction error of images in that class, a high reconstruction error value indicates images outside the class and thus, we compare the reconstruction error with a threshold to decide whether the image belongs to the class. 

\subsection{The proposed autoencoder}

\begin{figure*}
    \centering
    \includegraphics[width=0.9\textwidth]{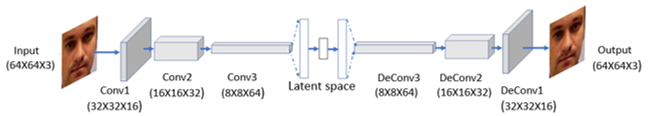}
    \caption{The architecture of the proposed convolutional autoencoder.}
    \label{fig:autoEncoder}
\end{figure*}

Figure~\ref{fig:autoEncoder} shows the architecture of the proposed autoencoder \cite{nikisins2019domain}. 
The input is a $64 \times 64$ RGB image; the encoder consists of three convolutions layers while the decoder, which is symmetric to the encoder and consists of three up-sampling layers, reconstructs a representation of original input image. We used nonlinear activation functions (RELU). 
 
All code was written in Python, on the Keras deep learning platform, and the experiments ran on an Intel Core i7 CPU64 GB RAM PC. The CAE network was trained with the RMSprop optimizer for 50 epochs, with a learning rate of 0.001. The batch 
the size was set to 32.

% ****************************************************************************** 
% ****************************************************************************** 
% ******************************************************************************

\subsection{Training, validation, and test datasets}

Faces on the images of all test and training datasets were detected with the Haar feature-based cascade classifier that was proposed by Viola and Jones in \cite{viola2001rapid}, followed by manual inspection and selection. User input was required, especially in the creation of training images from the in-the-wild, due to performance issues of the face detector on such images; image quality issues such as out of focus blurry faces; and in some cases the need to exclude imposter images, e.g. faces on a poster on a wall. All selected faces were cropped and normalized to $64 \times 64$ pixels.

We tested the autoencoder on two test datasets, the first from the Replay-Attack database and the other from the NUAA, consisting of 236 images each. The imposter part of the test datasets contained images from all types of attacks supported by these two databases. 

As the architecture and the training protocol of the autoencoder are fixed, the main variable of our experiment is the training set. As our aim is to see how the 
choice of training set affects the generalisation power of the classifier across the two test databases, we opted for training datasets such that D1 is subset of D2 and D3, and D2 subset of D3:  

\begin{itemize}
  \item[{\bf D1}] Images form the Replay-Attack data set only. We used 10 client subjects’ videos, both controlled and adverse. 
  \item[{\bf D2}] We added 163 face images harvested online using general keywords such as {\em teachers}. These 163 face images were manually chosen from a larger collection, the main considerations being to be frontal face images, in-focus, and of a good size so the normalisation to size $64 \times 64$ does not require excessive zooming. 
  \item[{\bf D3}] We added 269 images from the SCface and the CASIA-WebFace databases. The SCface is a surveillance camera face database from which we used the mugshot, still color images, captured indoors under controlled illumination conditions. The CASIA-WebFace is a very large dataset, consisting of 10,575 subjects, collected in a semi-automatic way from the Internet. We used a random subset of it.
 \end{itemize}
The description of the training datasets is summarized in Table~\ref{tbl:trainingSets}. 
\begin{table}
\centering
\begin{tabular}{|c|l|c|} \hline 
& \textbf{Description}& \textbf{size} \\ \hline 
{\bf D1} & Replay-Attack & 1027  \\ \hline 
{\bf D2} & Replay-Attack +Wild images  & 1190 \\ \hline
{\bf D3} & Combine Replay-Attack with others DB & 1459\\ \hline
\end{tabular}\newline
\caption{Description and size of the training datasets.}
\label{tbl:trainingSets}
\end{table}

%***************************************************

The validation dataset was kept constant to simplify the design of the experiment. It consisted of 208 images from the Replay-Attack, CASIA-Webface and in-the-wild images. As the use of a validation set with a composition similar to the training dataset {\bf D3} may lead to an underestimate of the performance of the performance of proposed autoencoder on Replay-Attack under intra-database protocol, we also report HTER values computed under the use of a validation set consisting of Replay-Attack images only.

% ****************************************************************************** 
% ****************************************************************************** 
% ****************************************************************************** 

\section{Results and discussion}

%\subsection{ protocols}
%For face anti-spoofing, the adaptation ability from one dataset to another is crucial for practical application. In 
%this part, we evaluate this ability by cross-dataset testing (intertest), namely the model is first trained using 
%the samples from dataset A (source), and then tested on dataset B (target).In order to assess the performance of our
%method, we proceed with the experiments by using two different protocols. The first one, consists in evaluates proposed 
%method inside the same antispoofing dataset, which is commonly known as intra-dataset
%evaluation. The second one was conducted in order to address
%the efficacy of our method when tested on another dataset, commonly being referred to as inter-dataset or cross-dataset evaluation. This latter one is the most challenging in the literature,
%due to the differences in scenery that one dataset shows from another one. we used for this for this Leave one dataset out protocol. 

%\subsection{Cross-database Testing}
%this is the most common protocol used to assess generalization of face-PAD algorithms. It is based on training on one or several data sets and testing in others. 

%\subsection{ Performance metrics}

\begin{figure*}[ht]
    \centering
        \includegraphics[width=0.475\linewidth]{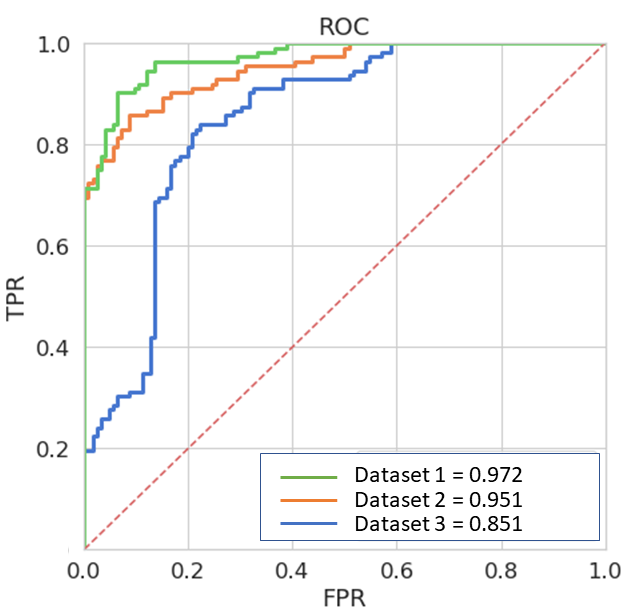} \hfill 
        \includegraphics[width=0.475\linewidth]{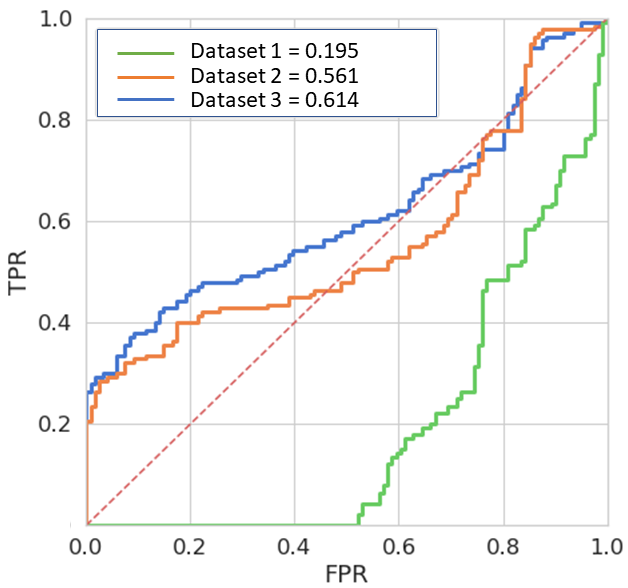}
    \caption{ROC curves corresponding to classifier/training dataset combinations, tested on Replay-Attack (top) and NUAA (bottom).}
    \label{fig:ROCs}
\end{figure*}

Figure~\ref{fig:ROCs} shows the ROC curves of the proposed autoencoder, trained on the three datasets, and tested on Replay-Attack (top) and NUAA (bottom). The corresponding Areas Under the Curve are reported on Table~2. We notice that the inclusion of the in-the-wild images in the training dataset improved markedly the cross-database generalisation power of the classifier with the value of the AUC on the NUAA going up from 0.19 to 0.56. Moreover, the inclusion of images from non-specialized databases further increased the AUC to 0.61. We also notice a moderate fall on the performance on Replay-Attack, with the AUC going down from .97 to .95 and then to .85. We also note the high performance of the algorithm under an intra-database test mode, that is, the high AUC value of .97 AUC on the Replay-Attack.  

\begin{table}[ht]
\begin{center}
\begin{tabular}{|r||c|c|c|} \hline 
& {\bf D1} & {\bf D2} & {\bf D3} \\ \hline 
Replay-Attack & .97  & .95 & .85  \\ \hline 
NUAA & .19 & .56 & .61 \\ \hline
\end{tabular} 
\caption{The AUC values corresponding to the ROC curves shown in Figure~\ref{fig:ROCs}.}
\end{center}
\label{tbl:AUCs}
\end{table}

% ***************************************************************

The value of the AUC is an integral over all possible operating points, that is, overall possible thresholds against which we compare the reconstruction error and determine whether a sample should be classified as client or imposter. Thus, it separates the problem of assessing the discriminative power of the classifier from the problem of finding an optimal, for the given test, operating point. Next, we will assess the proposed method on its capability to determine an optimal operating point. 

In the literature, classifier performance on a specific operating point is usually assessed either by reporting separately the False Positive Rate (FPR) and the False Negative Rate (FNR), or their mean average Half Total Error Rate. We note that reporting an operating specific performance metric does not necessarily mean that the problem of finding the optimal operating point has been addressed. For example, it could be reported the minimum HTER overall operating points, or the True Positive Rate corresponding to certain fixed values of FPR. Employing a commonly used technique to address this problem, we first compute a threshold on the validation set, here the threshold corresponding to the Equal Error Rate (EER) on that set, and use this threshold to compute the HTER. 

Table 3 summarizes the HTERs of our method and the corresponding HTERs reported in \cite{Rodrigo2019}. The high HTER values on NUAA, as opposed to the more satisfactory discriminating power of the classifier shown in the ROC curves, indicate that the threshold computed on the validation set cannot be used on NUAA. We note that \cite{Rodrigo2019} also reports a very high HTER, which again indicates that a satisfactory operating point on NUAA could not be found. Finally, we note that our convolutional autoencoder performs worse than \cite{Rodrigo2019} on Replay-Attack on intra-database test mode. However, we also note that our HTER value of .19 on Replay-Attack goes down to .15 when the non-Replay-Attack images are removed from the validation set. Moreover, the HTERs vales of 0.05 and 0.51 that we report in Table 3 for brevity, correspond to the lowest rates achieved by four different classifiers, which in \cite{Rodrigo2019} are reported to range from 0.05 to 0.32 for Replay-Attack and from .51 to .65 for NUAA.

\begin{table}[h]
\begin{center}
\begin{tabular}{|l|c|c|c|c|} \hline 
& {\bf D1} & {\bf D2} & {\bf D3} & \cite{Rodrigo2019} \\ \hline 
Replay-Attack  & .19 & .16 & .21 & .05 \\ \hline 
NUAA & .50 & .50 & .50 & .51 \\ \hline 
\end{tabular}\newline
\caption{HTERs computed on the operating point corresponding to the EER on the validation set. The last column shows the HTERs reported in \cite{Rodrigo2019}.}
\end{center}
\label{tbl:HTERs}
\end{table}

% ****************************************************************************** 
% ****************************************************************************** 
% ******************************************************************************

\section{Conclusions}

Our experiments show that one-class classifiers trained with images in-the-wild are a
promising research direction towards the development of face anti-spoofing algorithms that would be able to operate in uncontrolled environments and detect previously unseen 
types of attacks. In particular, we showed that the inclusion of such images in the training set of a convolutional autoencoder, which was originally trained on the Replay-Attack database, increased markedly its performance on the unseen NUAA database, as shown by the ROC curves and the corresponding AUC values. 

On the other hand, we also note that the algorithm could not operate successfully on both Replay-Attack and NUAA on a single operating point. Thus, the question of identifying suitable operating points for the algorithm to be able to work successfully on cross-database mode is still open. 

In the future, we plan to test the effect of the augmentation of the training set with in-the-wild images on various other anomaly detection algorithms, and also test the use of larger sets of in-the-wild images. The main challenge in the construction of very large training sets of in-the-wild images is that the typical online searches return many images that have undergone significant, and unknown to us, processing. We believe that the use of automatic image quality assessment algorithms, such as \cite{ferzli2009no}, can facilitate such an image selection task.  

% ****************************************************************************** 
% ****************************************************************************** 
% ******************************************************************************

\bibliographystyle{ieee}
\bibliography{references}

\end{document}